# Event-based clinical findings extraction from radiology reports with pre-trained language model

Wilson Lau[1], Kevin Lybarger[1], PhD, Martin L. Gunn[2], MB ChB, Meliha Yetisgen[1], PhD

[1]Biomedical & Health Informatics, School of Medicine, University of Washington, Seattle, WA
[2]Department of Radiology, School of Medicine, University of Washington, Seattle, WA

**Abstract**

Radiology reports contain a diverse and rich set of clinical abnormalities documented by radiologists during their interpretation of the images. Comprehensive semantic representations of radiological findings would enable a wide range of secondary use applications to support diagnosis, triage, outcomes prediction, and clinical research. In this paper, we present a new corpus of radiology reports annotated with clinical findings. Our annotation schema captures detailed representations of pathologic findings that are observable on imaging ("lesions") and other types of clinical problems ("medical problems"). The schema used an event-based representation to capture fine-grained details, including assertion, anatomy, characteristics, size, count, etc. Our gold standard corpus contained a total of 500 annotated computed tomography (CT) reports. We extracted triggers and argument entities using two state-of-the-art deep learning architectures, including BERT. We then predicted the linkages between trigger and argument entities (referred to as argument roles) using a BERT-based relation extraction model. We achieved the best extraction performance using a BERT model pre-trained on 3 million radiology reports from our institution: 90.9%-93.4% F1 for finding triggers 72.0%-85.6% F1 for arguments roles. To assess model generalizability, we used an external validation set randomly sampled from the MIMIC Chest X-ray (MIMIC-CXR) database. The extraction performance on this validation set was 95.6% for finding triggers and 79.1%-89.7% for argument roles, demonstrating that the model generalized well to the cross-institutional data with a different imaging modality. We extracted the finding events from all the radiology reports in the MIMIC-CXR database and provided the extractions to the research community.

## 1. Introduction

Radiology reports are the principal means for communicating and documenting diagnostic imaging results. The reports contain a diverse and rich set of information, including radiologic findings, diagnoses, and recommendations for follow-up tests. While there has been some limited exploration of structured radiology reports that capture findings through semantic representations [1], radiologists' findings are predominantly captured through unstructured text. Natural language processing (NLP) can automatically convert unstructured text, including radiology reports, to a structured semantic representation [2]. Structured semantic representations of the findings in radiology reports could facilitate many secondary use applications, including clinical decision-support systems [3], diagnostic surveillance of medical problems [4], identification of patient cohorts with specific phenotypes [5], tracking follow-up recommendations [6], image retrieval and data-mining [7], and simplification of report language for patients [8]. Large-scale and real-time use of radiological finding information in these types of secondary use applications requires a detailed semantic representation of the findings that captures the most salient information. Since imaging tests are commonly used for cancer screening and diagnosis, semantic representations for findings associated with lesions and medical problems would be largely applicable to secondary use.

In this paper, we explored the extraction of comprehensive representations of clinical findings from radiology reports, including the creation of a novel annotation schema, annotation of a new clinical data set, and the development of state-of-the-art clinical finding extraction models. In our annotation schema, we categorized findings in radiology reports as *Lesion* findings and *Medical Problem* findings. A *Lesion* finding was defined as an abnormal space-occupying mass that was observable on the images. Lesions included primary tumors, metastases, benign tumors, abscesses, nodules, and other masses. A *Medical Problem* finding was a pathological process that was not a lesion, for example cirrhosis, air-trapping, atherosclerosis, and effusion. Each finding category was represented through fine-grained event-based annotations. We presented a new annotated corpus of 500 computed tomography (CT) reports from the University of Washington (UW). To extract the finding events, we developed a deep learning extraction framework that fine-tuned a single BERT [9] model. We explored different contextualized embeddings through pre-training on different text sources. To assess the generalizability of the event extraction model, we annotated a subset of the MIMC-CXR radiology reports [10]. The extraction model achieved comparable performance on the MIMIC-CXR and UW data sets, despite the differences between the data sets. We extracted the clinical findings from the entire MIMIC-CXR data set and made the extracted findings available to the research community[1]. We also made the annotation guidelines and event extraction framework available[2]. The extraction framework directly processes annotated event data from the BRAT annotation tool [11] and can be readily used for event extraction without any deep learning coding experience.

**1.1 Related work**

The development of NLP-based information extraction (IE) models that target important information in clinical text has increased in recent decades [12]. Radiology is a clinical domain where NLP approaches, including IE, have been extensively applied [2]. Radiological finding information can be extracted by using named entity recognition (NER) to identify fine-grained details, such as anatomy, size, characteristics and assertion, and subsequently linking related phenomena using relation extraction (RE). Several studies employed custom rule-based linguistic patterns to identify clinical finding observations in radiology reports, including appendicitis indication, anatomy and assertion [13], adrenal observations and modifiers [14], and osteoporosis fracture categories and modifiers [15]. Due to the heterogeneity of writing styles, ambiguity of abbreviations, and presence of "hedging" statements [16], engineering linguistic and semantic rules to extract information from radiology reports requires substantial effort and clinical expertise. Furthermore, rule-based approaches produce brittle extraction models that do not generalize well. One example is the MedLEE system developed by Columbia University which incorporated comprehensive syntactic and semantic grammars to extract information from chest radiograph reports [17]. The conceptual model comprised 350 semantic grammar rules, 1,720 single-word lexicons, and 1,400 multi-word phrases. Development of the MedLEE semantic grammars required half a person-year [18], [19]. Sevenster et al. used MedLEE to identify finding observation and body location entities and establish relationships between entities through relations. However, the major drawback was that the recall of overall extraction (entities and relations) was less than 46% due to the lack of comprehensive lexicons and grammatical rules [20].

---

[1] https://github.com/uw-bionlp/MIMIC-CXR_clinical_findings
[2] https://github.com/wilsonlau-uw/BERT-EE

To overcome the limitations of rule-based systems, more contemporary radiology extraction work used statistical machine learning approaches to extract finding information. There is a body of radiology IE work that utilized discrete modeling approaches. For example, Hassanpour et al. used conditional Markov and conditional random field (CRF) models to extract anatomy, observations, modifiers, uncertainty entities from a corpus of 150 reports [21]. Yim et al. employed maximum entropy models to extract relations between tumor references and attributes from radiology reports of hepatocellular carcinoma patients [22]. One challenge with statistical machine learning approaches is that manually engineered features are often tailored to solve a specific problem and are not easily adaptable to other domains.

Recent radiology extraction studies utilize neural networks, which offer improved modeling capacity, abstraction, and transfer learning than discrete modeling approaches. A commonly applied neural approach is the sequence-based recurrent neural network (RNN) model, which encodes sequences using an internal memory mechanism. The Bidirectional Long Short-term Memory (BiLSTM) network is a popular RNN variant, which captures long-range sequential dependencies in the forward and backward directions. Cornegruta et al. extracted 4 different entities (body location, clinical finding, descriptor and medical device) with an annotated corpus of 2,000 radiology reports using BiLSTM [23]. Steinkamp et al. extracted clinical finding observations and their relations to modifier entities, such as location, size and change over time using another RNN variant, the Gated Recurrent Unit [24].

Most state-of-the-art NLP classification work, including IE within the radiology domain, utilized pre-trained transformer models with over hundreds of millions of model parameters. The popular BERT [9] model offers several benefits over RNN variants, including the combination of self-supervised pre-training and sub-token representation. BERT learns word relationships through a masked language modeling task and learns sentence dependency by predicting whether two sentences are adjacent. This pre-training process allows the model to develop deep representation of words in context through layers of multi-head self-attention. BERT intrinsically attends to certain types of syntactic relations [25], and the dependency information can be leveraged to increase relation extraction performance [26], [27]. Provided that the model is sufficiently pre-trained on unlabeled data in the target domain, the expressive contextual representations of BERT can be transferred to specific prediction tasks, including IE, and achieve state-of-the-art performance. Sugimoto et al. extracted 7 different clinical entities from a corpus of 540 Japanese CT radiology reports by fine-tuning a pre-trained Japanese BERT model [28]. Other studies extracted breast imaging entities and relations from Chinese radiology reports [29], [30]. Datta et al. employed a similar BERT fine-tuning approach to extract relations for clinical finding with spatial indication, such as "within" or "near" [31].

We identified several gaps in prior work that limit the creation of comprehensive semantic representations of findings in radiology reports, including: (1) the limited scope of the annotation and extraction schemas, (2) the limited scope of diseases and anatomy explored, and (3) the lack of demonstrated generalizability. Findings in radiology reports can be relatively complex, and several attributes are often needed to fully capture all the finding information present (e.g., assertion, anatomy, size, and other characteristics) for meaningful secondary use. Many prior studies only focused on entity extraction, without identifying the relations between entities in order to fully represent the findings [14], [15], [21], [23], [28]. To address this gap, we introduced an event-based annotation schema that captured a majority of the finding information. Several studies focused on specific diseases and/or anatomical regions [13], [22], [29]–[31]. While this focus may improve performance for the target diseases and/or anatomy, it reduces the generalizability of the annotated data sets and extraction models. To address this gap, we created the first general-purpose gold standard annotated with event-based schema on *Lesion* and *Medical Problem* findings without disease or anatomy constraints. The gold standard contained randomly sampled 500 CT reports. In comparison to the reports in other imaging modalities, such as chest X-ray reports, CT reports covered a wide range of anatomy, medical problems, lesion types, lesion characteristics, and assertions. We trained and evaluated the event extraction framework on this gold standard of CT reports. No other previous studies evaluated the generalizability of extraction models across imaging modalities or institutions. To address this gap, we evaluated the extraction performance on an external validation set we created from chest X-ray reports from the publicly available MIMC-CXR data set.

## 2. Materials and Methods

### 2.1 Dataset and Annotation Schema

We used an existing clinical dataset of 706,908 computed tomography (CT) reports from the UW clinical repository from 2008-2018. We randomly sampled 500 CT reports from this dataset and annotated as our gold standard corpus. Retrospective review of this dataset was approved by the UW institutional review board, and the dataset was de-identified to preserve the privacy of the patients and ensure HIPAA compliance.

Our annotation schema is summarized in Table 1. We used an event-based representation to capture the details of two clinical finding types: *Lesion* and *Medical Problem*. Each event was characterized with a trigger and a set of connected arguments. The trigger was a required key phrase identifying the finding event, while the arguments provided fine-grained details about the event. The argument entities were linked to the corresponding triggers through argument roles, forming a detailed and nuanced semantic representation of the clinical findings. We defined two types of arguments: *span-only* and *span-with-value*. The annotation of *span-only* arguments included the selection of the relevant phrase, assignment of an argument type label, and connection to the trigger, similar to most event annotation work. The annotation of *span-with-value* arguments included the selection of the relevant phrase, assignment of an argument type label with an additional categorical label that captures the clinical meaning of the selected phrase, as well as connection to the trigger. The categorical labels normalized the contents of the annotated phrase, allowing the extracted information to more easily be incorporated into secondary use applications. For example, in the sentence "No traumatic abnormality in the abdomen or pelvis", annotating the text span "no" as *Medical-Assertion* would also include the assignment of the categorical label *absent*. Because the presence of a lesion or medical problem could be implied rather than explicit, *present* was the default categorical label for *Assertion*, unless the report clearly indicated that the *possible* or *absent* labels were applicable.

|  | Argument | Type | Categorical labels | Span examples |
|---|---|---|---|---|
| Lesion Finding | Lesion Description (Trigger) | span-only | - | "mass", "lesion", "nodule" |
|  | Anatomy | span-only | - | "left lower lobe" |
|  | Assertion | span-with-value | present (*default*), absent, possible | "no", "possible" |
|  | Characteristics | span-only | - | "hypodense", "septal" |
|  | Count | span-only | - | "2", "numerous", "multiple" |
|  | Size | span-only | - | "4.1 x 3.1 cm", "small" |
|  | Size Trend | span-with-value | new, increasing, decreasing, no-change | "stable", "unchanged" |
| Medical Problem Finding | Medical Problem (Trigger) | span-only | - | "atherosclerotic calcifications" |
|  | Anatomy | span-only | - | "abdominal aorta", "right kidney" |
|  | Assertion | span-with-value | present (*default*), absent, possible | "no", "possible" |

**Table 1.** Annotation schema of *Lesion* finding and *Medical Problem* finding.

Extraction of these findings was treated as a slot filling task by identifying the text spans that corresponded to the arguments (argument entities with roles) of the clinical finding events. Figure 1 presents example annotations for a *Lesion* event and a *Medical Problem* event. For *span-only* arguments, the slot values would be the identified text spans. For *span-with-value* arguments, the slot values would be the identified categorical labels, which capture the meaning of the annotated phrases. A finding event might include multiple arguments of the same type. For example, a medical problem could be linked to multiple anatomical locations, or a lesion could be described by multiple characteristics.

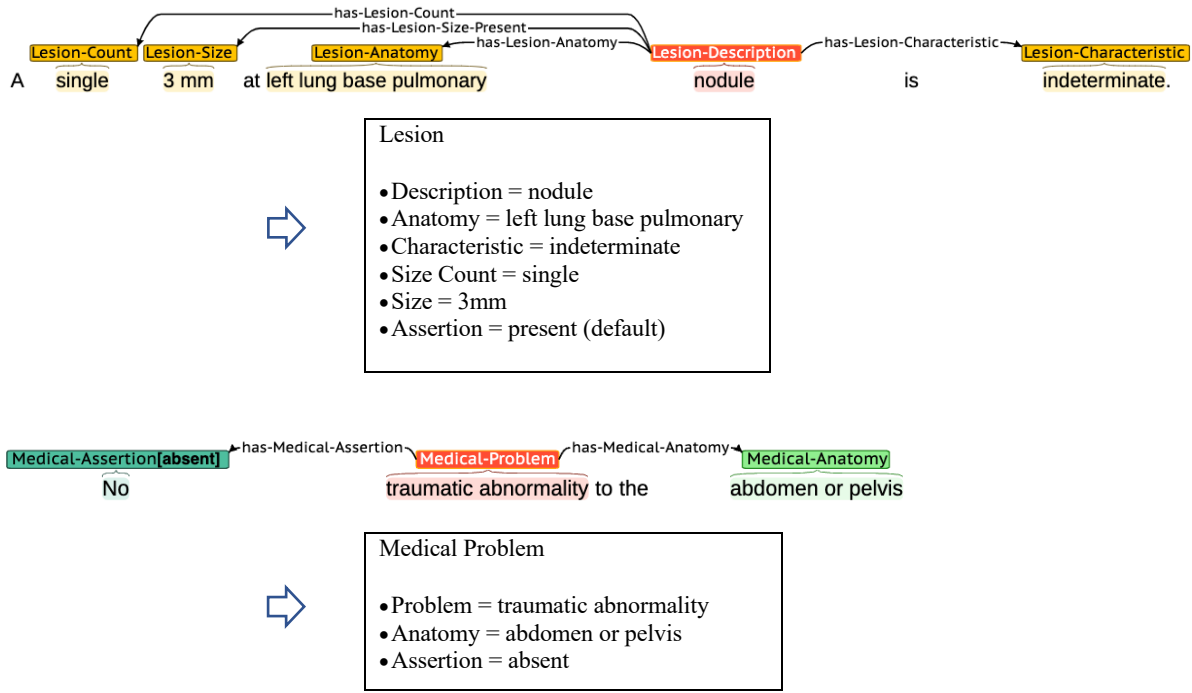

**Fig. 1** Example annotations for *Lesion* and *Medical Problem* events.

**2.2 Scoring criteria for evaluation**

Inter-annotator agreement and model extraction performance was evaluated using the same scoring criteria. The annotated and extracted events include trigger and argument entities that are connected through argument roles. The pairing of triggers and arguments (entities with identified roles) assembles events from the individual entities. The scoring criteria for trigger and argument entities and argument roles are presented below.

**2.2.1 Trigger and argument entities**

Trigger and argument entities scoring considered the span identification and labeling, without considering the roles linking trigger and argument entities. All trigger and argument entities were compared at the token-level (rather than span-level) to allow partial matches, since partially matched text spans could still contain clinically relevant information, e.g. "mass lesions" vs "lesions".

**2.2.2 Argument roles**

Argument role scoring considered three annotated/extracted phenomena: (1) the trigger entity, (2) the argument entity, and (3) the argument role (linking the trigger-argument entity pair). Argument role equivalence required the trigger entity, argument entity, and role label to be equivalent. In argument role scoring, the entity equivalence criteria for triggers, span-only arguments, and span-with-value arguments were based on the semantics of the event representation, by considering the most salient information being captured by the entities [32].

**Triggers**: Events were aligned based on trigger equivalence, and the arguments associated with aligned events (events with equivalent triggers) were compared based on the argument types. Triggers were considered equivalent if the spans overlapped by at least one token. Figure 2 shows an example of two *Medical Problem* annotations. Although the word "displaced" is not part of the trigger in Annotation #2, their overlapping text spans and connections to the *Medical-Anatomy* argument entities indicates that both argument entities belong to the same event and can be scored accordingly.

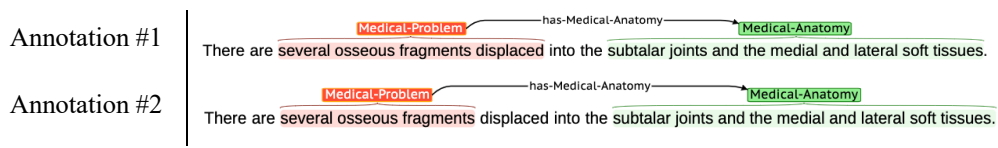

**Fig. 2** Two *Medical Problem* Finding event annotations with equivalent triggers.

**Span-only arguments**: When evaluating argument role performance, *span-only* argument entity equivalence was assessed at the token-level rather than span-level, because partial matches can capture clinically relevant information. The example in Figure 3, includes the same sentence with two sets of annotations for a *Lesion* event with multiple *Lesion-Anatomy* arguments. The second *Lesion-Anatomy* entities in the annotation do not match exactly. The discrepancy between the *Lesion-Anatomy* annotations ("extending" in Annotation #1) includes some clinical information; however, a majority of the clinically relevant information is captured by both spans ("posteriorly to the nasopharynx"). The token-level equivalence criteria for *span-only* argument entities was intended to reward such partial matches.

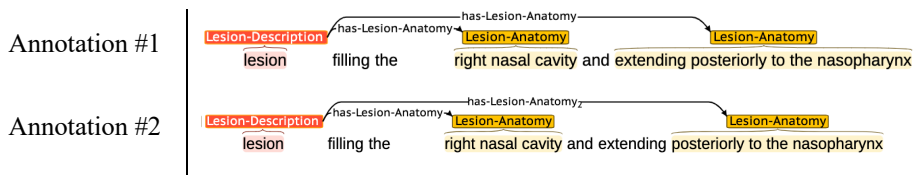

**Fig. 3** Two *Lesion Finding* event annotations with partially matched span-only arguments.

**Span-with-value arguments**: The categorical labels of *span-with-value* argument normalized the contents of the annotated phrase, allowing the extracted information to more easily be incorporated into secondary use applications. When evaluating argument role performance, the *span-with-value* argument entity equivalence was assessed based on the categorical labels only, without considering the spans. In Figure 4, although the *Lesion-Size-Trend* argument entity in Annotation #2 does not include the words "and number", both *Lesion-Size-Trend* annotations have the same categorical label and slot value (*increasing*). Hence both annotations are considered equivalent.

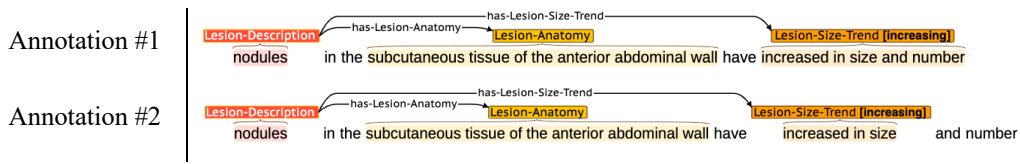

**Fig. 4** Two *Lesion Finding* event annotations with the same value for *Lesion-Size-Trend*.

**2.3 Gold standard corpus annotation**

The annotation was performed by one medical student and one graduate student using the BRAT rapid annotation tool [11]. Annotation guidelines were provided describing the details of each clinical finding event. In the initial iterations, the annotators were given the same samples to annotate independently. After each iteration, the annotators met with the domain expert radiologist to discuss and resolve the disagreements. The annotation guidelines were updated accordingly. At each iteration, we calculated inter-annotator agreement using pair-wise F1 score [33], by holding one set of annotated samples as gold standard and calculating the F1 of the other annotated samples. After four iterations, the final inter-annotator agreement over 30 CT reports was 93.0% F1 for triggers, 83.7% F1 for *span-only* arguments, and 86.9% F1 for *span-with-value* arguments, based on the argument role scoring in Section 2.2.2. The medical student single-annotated the remaining 470 CT reports. The final corpus contained 2,344 *Lesion* events (6,337 arguments entities and 6,617 argument roles) and 8,065 *Medical Problem* events (5,783 argument entities and 7,406 argument roles). The argument entity counts represented the number of annotated spans, and the argument role counts indicated the number of trigger-argument pairings. Since an argument entity could be linked to multiple triggers, the argument role counts could be greater than the argument entity counts. The distribution of the annotated argument entities and roles is shown in Table 2. As can be observed, the number of annotated *Medical Problem* events was more than three times higher than the number of *Lesion* events. In general, each argument type corresponded to a single argument role type (one-to-one mapping between argument types and roles). One exception is *Lesion-Size*, which could be connected to a trigger through a *Lesion-Size (Past)* or *Lesion-Size (Present)* argument role.

| Trigger/Argument entities | Frequency | Argument role | Frequency |
|---|---|---|---|
| Lesion-Description | 2,344 | - | |
| Lesion-Anatomy | 2,039 | Lesion-Anatomy | 2,187 |
| Lesion-Assertion | 945 | Lesion-Assertion | 1,008 |
| Lesion-Characteristic | 1,931 | Lesion-Characteristic | 1,968 |
| Lesion-Count | 235 | Lesion-Count | 237 |
| Lesion-Size | 816 | Lesion-Size (Past) | 94 |
|  |  | Lesion-Size (Present) | 736 |
| Lesion-Size-Trend | 371 | Lesion-Size-Trend | 387 |
| Medical-Problem | 8,065 | - | |
| Medical-Anatomy | 2,990 | Medical-Anatomy | 3,952 |
| Medical-Assertion | 2,793 | Medical-Assertion | 3,454 |

**Table 2.** Event annotation statistics.

Overall gold standard corpus statistics are presented in Table 3. On average, there were 16 *Medical Problem* events and 5 *Lesion* events annotated per report. Some CT reports in the gold standard were very dense and contained over 100 *Medical Problem* events.

|  | Minimum | Mean | Median | Maximum |
|---|---|---|---|---|
| Number of words per report | 50 | 327 | 288 | 1,383 |
| Number of events per report | 2 | 21 | 18 | 130 |
| Number of Medical Problem events per report | 0 | 16 | 13 | 129 |
| Number of Lesion events per report | 0 | 5 | 3 | 36 |
| Number of arguments per Medical Problem event | 0 | 1 | 1 | 5 |
| Number of arguments per Lesion event | 0 | 3 | 3 | 16 |

**Table 3.** Gold standard corpus statistics.

### 2.4 Event extraction

The finding events were extracted in two separate steps: (1) the trigger and argument entities were extracted and (2) the argument roles were identified by connecting extracted trigger and argument entities through relations. The pairing of the trigger and argument entities through the argument roles assembles events from the individual entity extractions. Our event extraction pipeline operated on sentences, which were treated as independent samples.

### 2.4.1 Trigger and argument entity extraction

The extraction of trigger and argument entities was defined as a NER task. For the *span-with-value* argument entities, the categorical labels were appended to in the entity labels, for example, *Medical-Assertion (absent)*. Predicting the labels of the argument entities would therefore predict both the argument type and the categorical labels. We evaluated two state-of-the-art neural network architectures: (1) BiLSTM-CRF [34] and (2) BERT NER [9]. BiLSTM-CRF was considered a strong NER baseline by multiple studies [28], [30], [31]. We used the open source NeuroNER [35] for the BiLSTM-CRF implementation. Figure 5 presents NeuroNER's BiLSTM-CRF architecture. Each token in the input sentence was represented by the concatenation of a pretrained word embedding and a character-aware word embedding. The character-aware word embedding was generated by a BiLSTM operating on the individual characters associated with each token. The character-aware word embedding enabled the model to learn the morphological structure in each word and to encode out-of-vocabulary tokens. The sequence of word embeddings was then encoded using a second BiLSTM layer to create a contextualized representation of the sentence. The label of each word was predicted by a CRF output layer which took into account the conditional dependencies across the neighboring labels. To create input labels for the NER model from our annotated corpus, we used the Begin, Inside, Outside (BIO) tagging schema, based on whether the token was at the beginning, inside or outside of a label. For instance, consider the sentence "*Probable malignant pancreatic mass with no evidence of vascular encasement*". The labels would be classified as illustrated in Figure 5.

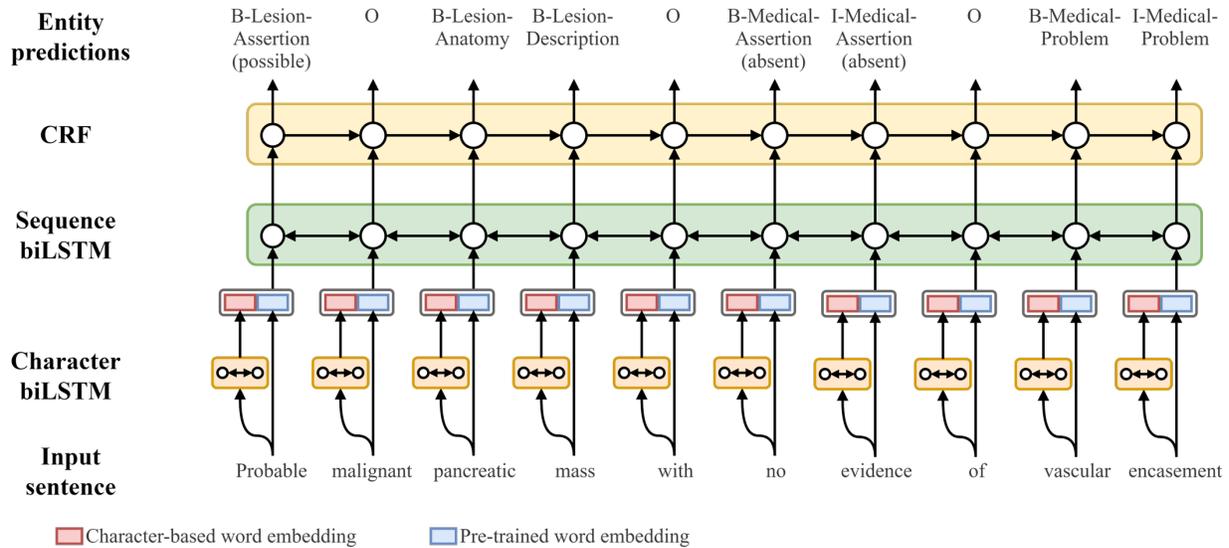

**Fig. 5** Architecture of the NeuroNER BiLSTM-CRF model.

The BERT NER model was implemented by adding a single linear layer to the BERT output hidden states and fine-tuning a pre-trained BERT model, as described by Devlin et al [9]. Because BERT utilized WordPiece tokenization [36], rare words would be segmented into multiple sub-tokens. These sub-tokens, prefixed by "##" if not the first sub-token, allowed the segments of the words to be represented in a deterministic fashion. Rather than using a universal token like [UNK], the sub-token representation provided richer contextual embeddings for the model to generalize. During the BIO labeling, the sub-tokens starting with "##" were assigned a special label #. In addition, the BERT input included the special tokens [CLS] and [SEP] at the beginning and end of a sentence respectively, to signify the sentence boundaries. Figure 6 illustrates how the labels of an input sentence were classified by BERT NER.

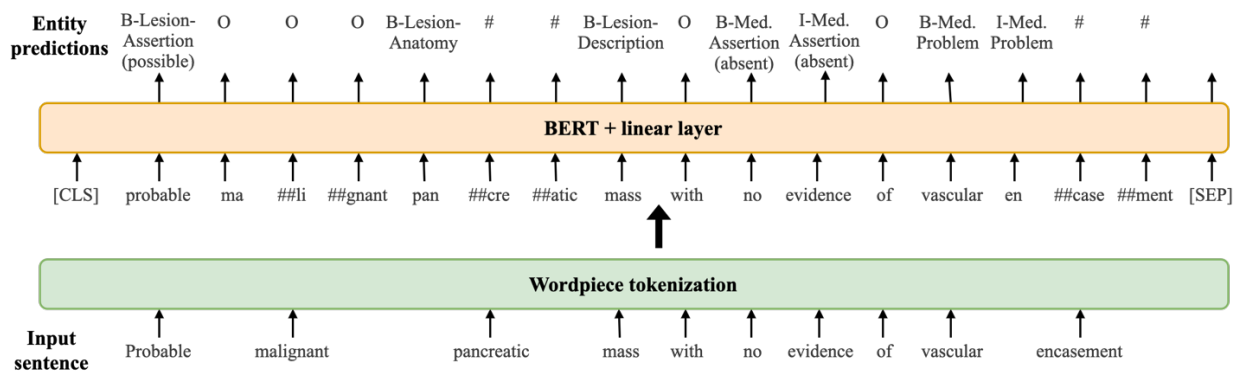

**Fig. 6** Architecture of the BERT NER model.

## 2.4.2 Argument role extraction

Once the trigger and argument entities were extracted, the argument roles were identified by predicting the links between trigger and argument entities. Identifying the roles of the argument entities filled the slots of the clinical finding events, similar to Figure 1. Each event included a trigger that anchored the event, with zero or more argument connections. Each argument role was represented by a unidirectional relation where the head was the trigger entity and the tail was an argument entity. We predicted the argument roles, by decomposing each event into a set of relations, predicting the relations, and then assembling events from the predicted relations.

Relations were extracted using BERT by adding a linear layer to the pooled output state (encoded in the [CLS] token) and fine-tuning the model. Figure 7 presents the BERT relation extraction (RE) model with an example input sentence. A unique input sentence was created for each candidate trigger-argument relation. The trigger and argument entity locations were marked with two pairs of special tokens, namely ([unused0], [unused1]) and ([unused2], [unused3]), which provided positional information about the entities and the direction of the relation (disambiguate head and tail). These special tokens were part of the BERT vocabulary designed for introducing new domain specific samples for fine tuning purposes. Consider the aforementioned example where the word "*Probable*" is the *Lesion-Assertion* of the *Lesion* trigger "*mass*". The trigger would be marked as "[unused0] mass [unused1]" and the *Lesion-Assertion* would be marked as "[unused2] Probable [unused3]". In addition, we introduced a new relation called "No_relation" for negative training instances indicating the absence of relations between some argument entities and triggers.

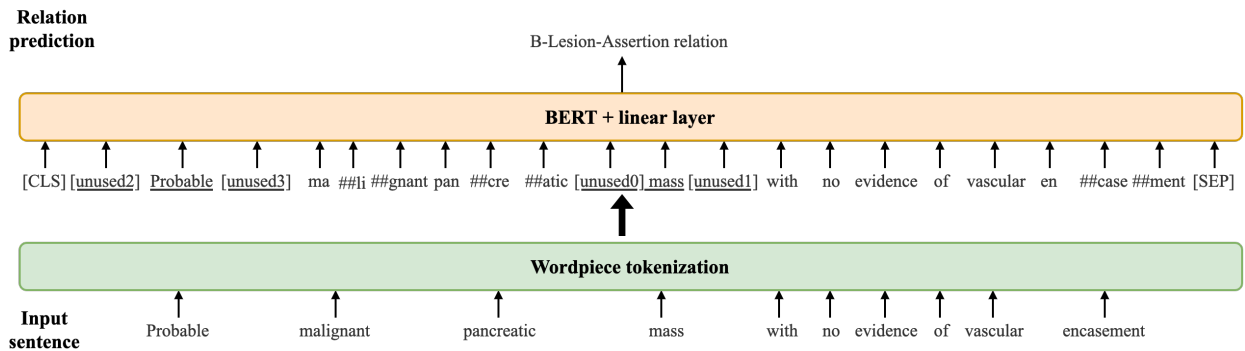

**Fig. 7** Architecture of the BERT RE model.

A single BERT model was fine-tuned for both the NER and RE tasks. In each training epoch, the NER and RE batches were alternated randomly, minimizing the cross-entropy loss for the applicable target (NER or RE), and thereby effectively allowing the model to learn from the two different tasks.

We performed 5-fold cross validation (CV) for all experiments using the same data split ratio (80% for training, 10% for validation, 10% for testing). The validation set was used for applying early stopping in order to avoid overfitting the training data [38]. The training was stopped when the validation results no longer showed improvement.

For the entity extraction baseline (BiLSTM-CRF$_{rad}$), we used the word2vec embeddings pre-trained on a radiology report dataset from our previous work [37]. This dataset contained over 3 million reports covering a wide range of imaging modalities and were collected from four institutions including the University of Washington Medical Center, Northwest Hospital and Medical Center, the Seattle Cancer Care Alliance, and Harborview Medical Center. In terms of the model hyperparameters, the embedding dimension and the hidden state dimension of the character and sequence LSTM layers were 25 and 100. We used the Adam Optimizer with a learning rate of 0.005, as suggested by NeuroNER.

We experimented with three different pre-trained BERT models (BERT$_{base}$, BERT$_{clinical}$ and BERT$_{rad}$). BERT$_{base}$ was pre-trained on Wikipedia and BookCorpus, and made available by Google [9]. BERT$_{clinical}$ was pre-trained on 2 million clinical notes, including over 500,000 radiology reports, from the MIMIC-III database [39], [40]. BERT$_{rad}$ was pre-trained on over 3 million UW radiology reports and was initialized from the BERT$_{clinical}$. We pre-trained BERT$_{rad}$ for 150,000 steps with a batch size of 32, maximum sequence length of 128, and a learning rate of 2e-5. In our experiments, both entities and relations were extracted by fine-tuning the same BERT model. We used the same set of hyperparameters in all the extraction experiments, based on the recommended values suggested by Devlin et al., with a learning rate of 3e-5, a drop-out rate of 0.1. Early stopping was also employed using the validation set.

To better assess the general performance of the models with different subsamples, we repeated the cross validation 10 times. For each run, the cross validation data splits were created with a different random seed [38]. We reported the average precision, recall and F1 scores across these 50 different runs and included the 95% confidence intervals.

## 3. Results

### 3.1 Trigger and argument entity extraction results

All of the trigger and argument entities were extracted first before their relations were identified. Trigger and argument entity extraction performance was evaluated at the token-level, as described in Section 2.2.1. The results are shown in Table 4.

| Entity | BiLSTM-CRF$_{rad}$ | | | BERT$_{base}$ | | | BERT$_{clinical}$ | | | BERT$_{rad}$ | | |
|---|---|---|---|---|---|---|---|---|---|---|---|---|
| | P | R | F1 | P | R | F1 | P | R | F1 | P | R | F1 |
| Medical-Problem | 88.8 | 84.9 | 86.7 (±0.45) | 89.1 | 83.9 | 86.4 (±0.37) | 90.5 | 83.6 | 86.8 (±0.37) | 91.3 | 85.0 | **88.0** (±0.34) |
| Medical-Anatomy | 79.1 | 79.9 | 79.3 (±0.92) | 82.3 | 77.9 | 79.9 (±0.87) | 83.8 | 77.3 | 80.3 (±0.84) | 85.7 | 78.5 | **81.8** (±0.75) |
| Medical-Assertion | 85.6 | 84.5 | 84.9 (±0.79) | 86.9 | 85.7 | 86.3 (±0.70) | 87.8 | 84.7 | 86.1 (±0.63) | 88.5 | 86.3 | **87.3** (±0.78) |
| Lesion-Description | 87.2 | 87.9 | 87.5 (±0.71) | 89.1 | 86.8 | 87.9 (±0.66) | 89.0 | 87.6 | 88.2 (±0.62) | 90.0 | 88.4 | **89.1** (±0.63) |
| Lesion-Anatomy | 80.2 | 78.6 | 79.0 (±0.92) | 85.5 | 76.5 | 80.6 (±0.94) | 85.8 | 76.8 | 80.8 (±0.89) | 86.8 | 80.7 | **83.5** (±0.86) |
| Lesion-Assertion | 81.3 | 72.1 | 76.2 (±1.55) | 86.0 | 70.0 | 76.8 (±1.60) | 85.6 | 70.5 | 77.1 (±1.48) | 86.5 | 73.6 | **79.3** (±1.26) |
| Lesion-Characteristic | 76.6 | 72.6 | 74.1 (±1.36) | 81.8 | 70.5 | 75.4 (±1.22) | 82.8 | 71.3 | 76.3 (±1.11) | 84.2 | 73.6 | **78.3** (±1.14) |
| Lesion-Size | 84.1 | 85.8 | 84.4 (±1.88) | 91.1 | 84.2 | 87.3 (±1.37) | 89.1 | 84.4 | 86.4 (±1.56) | 90.7 | 88.2 | **89.3** (±1.43) |
| Lesion-Count | 89.1 | 85.6 | 86.7 (±2.20) | 90.9 | 86.6 | 88.0 (±2.15) | 92.0 | 88.0 | **89.3** (±2.07) | 91.0 | 87.5 | 88.7 (±2.16) |
| Lesion-Size-Trend | 69.0 | 63.2 | 65.5 (±3.20) | 78.0 | 60.7 | 67.6 (±3.14) | 75.2 | 59.5 | 65.5 (±2.98) | 77.3 | 63.6 | **68.9** (±3.06) |
| Overall | 84.2 | 82.1 | 83.1 (±0.37) | 86.7 | 80.9 | 83.7 (±0.36) | 87.7 | 80.6 | 84.0 (±0.28) | 88.8 | 82.4 | **85.5** (±0.28) |

**Table 4.** Entity extraction results (average precision, recall and F1 in %), based on 10 runs of 5-fold cross validation. The numbers in brackets are 95% confidence intervals of the averages. The best F1 values are in bold.

All of the BERT implementations outperformed BiLSTM-CRF$_{rad}$. The BERT model with radiology-specific pretraining, BERT$_{rad}$, generally performed better than the other variants, BERT$_{base}$ and BERT$_{clinical}$, achieving the highest overall average F1 of 85.5%. In *Lesion-Count* prediction, BERT$_{clinical}$ is slightly higher than BERT$_{rad}$. In *Lesion-Size-Trend* prediction, the *decreasing* label had relatively low extraction performance due to the small sample size. For the *Assertion* extraction, the *absent* label was easier to predict since most of the annotated text spans comprised a single word "no", which constituted 70% of the *Medical-Assertion* and 84% of the *Lesion-Assertion* entities.

We conducted statistical significance tests using the overall F1 to access whether the difference in model results were due to randomness or sampling variability. In cross validation, the training sets overlap between different folds. As a result, the classification performance from each fold is not completely independent, and can lead to misleading statistical results when applying standard paired t-tests [41]. Hence, we applied the corrected resampled t-test, as suggested by Nadeau and Bengio [42], to better estimate the sample variance. The test results showed that the overall performance of BERT$_{rad}$ was better than the other architectures with significance (p-value < 5e-6).

### 3.2 Argument role extraction results

In this section, we present the end-to-end argument role extraction results. Specifically, we predicted the argument roles using the extracted triggers and argument entities rather than the gold standard entities. Table 5 shows the extraction results based on the scoring criteria described in Section 2.2.2. In general, the in-domain contextualized representations helped the BERT$_{rad}$ model achieved higher performance (except *Lesion-Count*).

| Argument type | Argument role | BERT$_{base}$ | | | BERT$_{clinical}$ | | | BERT$_{rad}$ | | |
|---|---|---|---|---|---|---|---|---|---|---|
| | | P | R | F1 | P | R | F1 | P | R | F1 |
| Span-only | Medical-Anatomy | 78.4 | 67.1 | 72.1 (±1.12) | 80.0 | 66.6 | 72.5 (±1.02) | 81.4 | 68.3 | **74.2** (±1.00) |
| Span-with-value | Medical-Assertion | 86.8 | 82.3 | 84.5 (±0.54) | 87.5 | 81.7 | 84.4 (±0.43) | 88.6 | 83.0 | **85.6** (±0.45) |
| Span-only | Lesion-Anatomy | 83.6 | 67.7 | 74.7 (±1.15) | 84.2 | 68.1 | 75.1 (±0.98) | 84.7 | 71.3 | **77.3** (±1.06) |
| | Lesion-Characteristic | 80.4 | 65.2 | 71.6 (±1.32) | 81.5 | 66.0 | 72.6 (±1.21) | 82.6 | 67.9 | **74.2** (±1.28) |
| | Lesion-Count | 87.0 | 81.6 | 83.4 (±2.11) | 89.8 | 83.6 | **86.0** (±2.18) | 88.1 | 83.3 | 85.1 (±2.09) |
| | Lesion-Size | 85.1 | 59.9 | 69.9 (±2.56) | 85.5 | 60.6 | 70.5 (±2.10) | 86.4 | 62.5 | **72.0** (±2.25) |
| Span-with-value | Lesion-Assertion | 85.4 | 79.7 | 82.4 (±0.69) | 84.9 | 80.0 | 82.3 (±0.76) | 86.1 | 81.2 | **83.5** (±0.61) |
| | Lesion-Size-Trend | 82.1 | 71.4 | 76.0 (±1.94) | 80.3 | 70.4 | 74.4 (±2.21) | 81.9 | 74.1 | **77.4** (±2.28) |

**Table 5.** End-to-end arguments role extraction results (average precision, recall and F1 in %), based on 10 runs of 5-fold cross validation. The numbers in brackets are 95% confidence intervals of the averages. The best F1 values are in bold.

### 3.3 Overall trigger and argument role extraction results

Table 6 presents the overall trigger and argument role extraction performance, evaluated with the scoring described in section 2.2.2. The BERT$_{rad}$ model achieved the highest average F1 of 92.9% for triggers (93.4% for *Medical-Problem* and 90.0% for *Lesion-Description*), suggesting very high overlap between the extracted findings and the gold standard. The highest average F1 for *span-only* arguments and *span-with-value* arguments were 75.0% and 84.8% respectively. The performance of BERT$_{base}$ was comparable to BERT$_{clinical}$. While BERT$_{clinical}$ performed slightly better than BERT$_{base}$ on triggers and *span-only* arguments, BERT$_{base}$ performed slightly better on *span-with-value* arguments.

| Argument type | BERT$_{base}$ | | | BERT$_{clinical}$ | | | BERT$_{rad}$ | | |
|---|---|---|---|---|---|---|---|---|---|
| | P | R | F1 | P | R | F1 | P | R | F1 |
| Trigger | 90.9 | 92.1 | 91.5 (±0.24) | 91.5 | 92.2 | 91.8 (±0.26) | 92.6 | 93.2 | **92.9** (±0.25) |
| Span-only | 79.8 | 67.1 | 72.8 (±0.71) | 81.1 | 67.0 | 73.3 (±0.66) | 82.3 | 69.0 | **75.0** (±0.66) |
| Span-with-value | 86.3 | 81.2 | 83.6 (±0.46) | 86.3 | 76.3 | 83.5 (±0.41) | 87.6 | 82.1 | **84.8** (±0.39) |

**Table 6.** Overall extraction performance for each type of arguments (average Precision, Recall and F1 in %).

We conducted the same statistical tests on the event argument extraction results using the overall performance scores presented in Table 6. BERT$_{rad}$ achieved the best overall performance with significance (p-values < 1.6e-4).

### 4. Extracting findings from MIMIC-CXR radiology reports

We used the chest X-ray reports in the MIMC-CXR database, to explore the generalizability of the event extraction model that was trained on CT reports. The MIMIC-CXR database consists of 337,110 images in 227,835 radiographic studies performed at the emergency department of the Beth Israel Deaconess Medical Center from 2011-2016. Each study is associated with a single radiology report [10]. The dataset was made publicly accessible to support independent research, such as predicting pulmonary edema severity [43], predicting COVID-19 pneumonia severity [44], and evaluating FDA approved AI devices [45].

To evaluate the generalizability of our extraction model, we manually annotated 50 randomly selected chest X-ray reports from the MIMC-CXR database using the same finding event annotation schema. This validation set included 257 *Medical Problem* finding events (141 argument entities and 313 roles) and 7 *Lesion* finding events (9 argument entities, 15 roles). The overall F1 scores on this validation set were 95.6% for triggers, 79.1% for *span-only* arguments and 89.7% for *span-with-value* arguments, evaluated using the same argument role scoring criteria in Section 2.2.2. The extraction performance was comparable to our repeated 5-fold cross validation performance, despite the fact that the MIMC-CXR reports were from a different institution and based on a different imaging modality. The MIMC-CXR radiology reports were generally shorter than the reports in our training corpus. The statistics of word count per report had a mean of 87, and a median of 79, in comparison to the mean and median of 327 and 288 in our corpus. We found that the event extraction model was able to identify clinical concepts that were unseen in our training corpus. For instance, the words "plasmacytoma" and "fibroadenomas" were correctly identified as lesions and "acute respiratory distress syndrome" was correctly identified as medical problem, even though these lesion and medical problem mentions did not appear in any radiology reports in the training corpus. This could be attributed to the pre-training of BERT$_{rad}$ with 3 million UW radiology reports covering a wide range of modalities.

We extracted lesion and medical problem findings from all 227,835 chest X-ray reports in the MIMIC-CXR dataset with our event extraction framework. A total of 1,420,604 medical problem findings and 31,706 lesion findings were extracted using the fine-tuned BERT$_{rad}$ model. To contribute to the core aim of the MIMIC-CXR project and facilitate future research studies in medical imaging, we are releasing the finding extraction results for all 227,835 radiology reports. The extracted data are in BRAT's standoff format and follow the same subject IDs, study IDs and folder structure, such that they can be readily used to augment the existing images and reports[3].

## 5. Discussion

We presented a new schema for representing lesion and medical problem findings in radiology reports. In trigger and argument entity extraction, the BERT-based NER models outperformed the BiLSTM-CRF baseline. In both the entity extraction and argument role prediction tasks, the BERT model with the most domain-specific pre-training, $BERT_{rad}$, achieved the best performance. Pre-training $BERT_{rad}$ on 3 million UW radiology reports allowed the model to learn better contextual representations and transfer the knowledge of clinical concepts that are absent in the training corpus. $BERT_{rad}$ achieved an end-to-end performance of 92.9% F1 for triggers, 75.0% F1 for *span-only* arguments, and 84.8% F1 for *span-with-value* arguments.

Among the finding entities, *Medical-Problem* and *Medical-Anatomy* had relatively long text spans. Over 25% of *Medical-Problem* spans and 35% of *Medical-Anatomy* spans contained at least 5 words. We found that some entities with lengthy spans were extracted into multiple separate entities, particularly before and after a conjunctive word. About 4% of all *Medical-Problem* entities and 7% of all *Medical-Anatomy* entities were split into multiple entities by the entity extraction models. Figure 8 presents an example of each case.

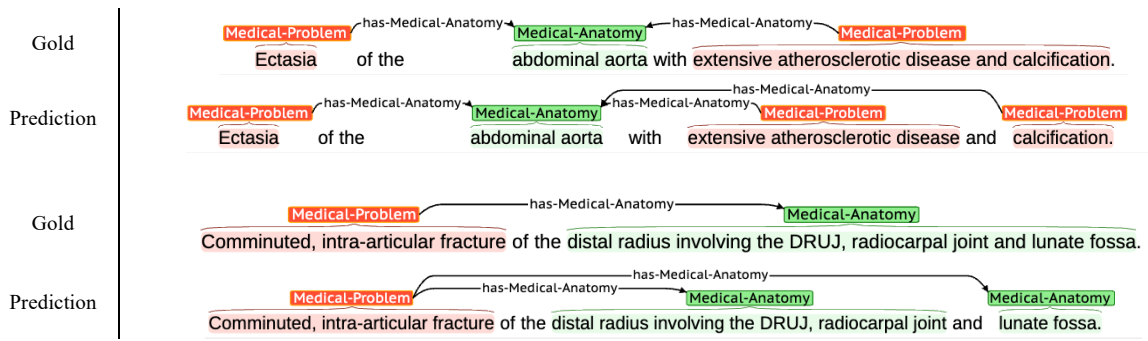

**Fig. 8** Examples of long text spans being extracted into multiple entities.

In our annotation schema, the same entity could be assigned multiple labels. For example, the same anatomy span could possibly be annotated as both *Lesion-Anatomy* and *Medical-Anatomy*. Our NER models could only assign a single label to each token, so a text span cannot be extracted as multiple argument entities. Approximately 1% of all entities in our annotated corpus had multiple labels, so this limitation does not fundamentally impact extraction performance. One way to circumvent this single-label limitation is by having a single entity for both findings.

---
[3] https://physionet.org/content/mimic-cxr/2.0.0/

Although a single anatomy entity no longer carries any clinical finding connotation, its association with the finding events can still be identified by the RE model.

Our extraction framework employed multi-task learning to optimize the parameters of a single BERT model. We did not explore other fine-tuning approaches, such as using graph structures to jointly model the span relations in the different tasks [46] or using entity aware markers to encode input sentences in a relation extraction model, which was shown to outperform joint modeling architectures [47]. Our BERT$_{rad}$ model was pre-trained using the common transfer learning paradigm by initializing its weight from another BERT model in relevant domain. This approach is particularly advantageous when the target data is scarce. However, a recent study showed that pre-training the language model from scratch in a domain with abundant unlabeled text could derive better in-domain vocabulary and result in substantial performance improvement [48]. Since our UW data set contained more than 3 million radiology reports, this pre-training approach could potentially improve the contextual representation of the BERT$_{rad}$ model and possibly lead to better event extraction performance.

## 6. Conclusion

In this paper, we presented a new schema for extracting lesion and medical problem findings from radiology reports. The event representation of each clinical finding comprised a trigger and different arguments, capturing the fine-grained semantic information of the finding. A total of 2,344 lesion findings and 8,065 medical problem findings were annotated in 500 CT radiology reports. For argument entity extraction, we evaluated two state-of-the-art neural architectures using BiLSTM-CRF and BERT. The BERT$_{rad}$ model pre-trained on 3 million radiology reports achieved an overall average F1 score of 85.5%, based on token-level evaluation. We then extracted the clinical finding events by predicting the argument roles for the extracted entities. The overall average F1 scores for end-to-end event extraction were 92.9% for triggers, 75.0% for *span-only* arguments and 84.8% for *span-with-value* arguments. To demonstrate the generalizability of the BERT$_{rad}$ model, we extracted the clinical findings (1,420,604 medical problem findings and 31,706 lesion findings) from all the radiology reports in the MIMIC-CXR database. Based on the evaluation with a manually labeled validation set of 50 chest X-ray reports, the overall average F1 scores for the extraction were 95.6% for triggers, 79.1% for *span-only* arguments and 89.7% for *span-with-value* arguments. The extraction performance was comparable to the repeated 5-fold cross validation performance with the UW corpus. We are releasing both our deep learning event extraction framework as well as the MIMIC-CXR extracted clinical findings to the research community.

## 7. Acknowledgements

This publication was supported by NIH/NCI (1R01CA248422-01A1), NIH/NLM Biomedical and Health Informatics Training Program (5T15LM007442-19), NIH/NCATS (UL1 TR002319). Research and results reported in this publication was facilitated by the generous contribution of computational resources from the University of Washington Department of Radiology.

**References**

[1] D. L. Rubin and C. E. Kahn, "Common Data Elements in Radiology," *Radiology*, vol. 283, no. 3, pp. 837–844, Jun. 2017, doi: 10.1148/radiol.2016161553.

[2] E. Pons, L. M. M. Braun, M. G. M. Hunink, and J. A. Kors, "Natural Language Processing in Radiology: A Systematic Review," *Radiology*, vol. 279, no. 2, pp. 329–343, Apr. 2016, doi: 10.1148/radiol.16142770.

[3] D. Demner-Fushman, W. W. Chapman, and C. J. McDonald, "What can natural language processing do for clinical decision support?," *Journal of Biomedical Informatics*, vol. 42, no. 5, pp. 760–772, Oct. 2009, doi: 10.1016/j.jbi.2009.08.007.

[4] J. P. Haas, E. A. Mendonça, B. Ross, C. Friedman, and E. Larson, "Use of computerized surveillance to detect nosocomial pneumonia in neonatal intensive care unit patients," *American Journal of Infection Control*, vol. 33, no. 8, pp. 439–443, Oct. 2005, doi: 10.1016/j.ajic.2005.06.008.

[5] K. N. Danforth, M. I. Early, S. Ngan, A. E. Kosco, C. Zheng, and M. K. Gould, "Automated Identification of Patients With Pulmonary Nodules in an Integrated Health System Using Administrative Health Plan Data, Radiology Reports, and Natural Language Processing," *Journal of Thoracic Oncology*, vol. 7, no. 8, pp. 1257–1262, Aug. 2012, doi: 10.1097/JTO.0b013e31825bd9f5.

[6] T. Mabotuwana *et al.*, "Automated Tracking of Follow-Up Imaging Recommendations," *American Journal of Roentgenology*, vol. 212, no. 6, pp. 1287–1294, Jun. 2019, doi: 10.2214/AJR.18.20586.

[7] A. Gerstmair, P. Daumke, K. Simon, M. Langer, and E. Kotter, "Intelligent image retrieval based on radiology reports," *Eur Radiol*, vol. 22, no. 12, pp. 2750–2758, Dec. 2012, doi: 10.1007/s00330-012-2608-x.

[8] B. Qenam, T. Y. Kim, M. J. Carroll, and M. Hogarth, "Text Simplification Using Consumer Health Vocabulary to Generate Patient-Centered Radiology Reporting: Translation and Evaluation," *J Med Internet Res*, vol. 19, no. 12, p. e417, Dec. 2017, doi: 10.2196/jmir.8536.

[9] J. Devlin, M.-W. Chang, K. Lee, and K. Toutanova, "BERT: Pre-training of Deep Bidirectional Transformers for Language Understanding," in *Proceedings of the 2019 Conference of the North American Chapter of the Association for Computational Linguistics: Human Language Technologies, Volume 1 (Long and Short Papers)*, Minneapolis, Minnesota, Jun. 2019, pp. 4171–4186. doi: 10.18653/v1/N19-1423.

[10] A. E. W. Johnson *et al.*, "MIMIC-CXR, a de-identified publicly available database of chest radiographs with free-text reports," *Sci Data*, vol. 6, no. 1, p. 317, Dec. 2019, doi: 10.1038/s41597-019-0322-0.

[11] P. Stenetorp, S. Pyysalo, G. Topić, T. Ohta, S. Ananiadou, and J. Tsujii, "brat: a Web-based Tool for NLP-Assisted Text Annotation," in *Proceedings of the Demonstrations at the 13th Conference of the European Chapter of the Association for Computational Linguistics*, Avignon, France, Apr. 2012, pp. 102–107. Accessed: Jan. 02, 2020. [Online]. Available: https://www.aclweb.org/anthology/E12-2021

[12] Y. Wang *et al.*, "Clinical information extraction applications: A literature review," *Journal of Biomedical Informatics*, vol. 77, pp. 34–49, Jan. 2018, doi: 10.1016/j.jbi.2017.11.011.

[13] B. Rink *et al.*, "Extracting Actionable Findings of Appendicitis from Radiology Reports Using Natural Language Processing," *AMIA Jt Summits Transl Sci Proc*, vol. 2013, p. 221, 18 2013.

[14] J. J. Zopf, J. M. Langer, W. W. Boonn, W. Kim, and H. M. Zafar, "Development of Automated Detection of Radiology Reports Citing Adrenal Findings," *J Digit Imaging*, vol. 25, no. 1, pp. 43–49, Feb. 2012, doi: 10.1007/s10278-011-9425-7.

[15] Y. Wang, S. Mehrabi, S. Sohn, E. J. Atkinson, S. Amin, and H. Liu, "Natural language processing of radiology reports for identification of skeletal site-specific fractures," *BMC Med Inform Decis Mak*, vol. 19, no. Suppl 3, p. 73, Apr. 2019, doi: 10.1186/s12911-019-0780-5.

[16] F. M. Hall, "Language of the radiology report: primer for residents and wayward radiologists," *AJR Am J Roentgenol*, vol. 175, no. 5, pp. 1239–1242, Nov. 2000, doi: 10.2214/ajr.175.5.1751239.

[17] C. Friedman, S. B. Johnson, B. Forman, and J. Starren, "Architectural requirements for a multipurpose natural language processor in the clinical environment.," *Proc Annu Symp Comput Appl Med Care*, pp. 347–351, 1995.

[18] C. Friedman, J. J. Cimino, and S. B. Johnson, "A conceptual model for clinical radiology reports.," *Proc Annu Symp Comput Appl Med Care*, pp. 829–833, 1993.

[19] C. Friedman, P. O. Alderson, J. H. Austin, J. J. Cimino, and S. B. Johnson, "A general natural-language text processor for clinical radiology.," *J Am Med Inform Assoc*, vol. 1, no. 2, pp. 161–174, 1994.

[20] M. Sevenster, R. van Ommering, and Y. Qian, "Automatically Correlating Clinical Findings and Body Locations in Radiology Reports Using MedLEE," *J Digit Imaging*, vol. 25, no. 2, pp. 240–249, Apr. 2012, doi: 10.1007/s10278-011-9411-0.

[21] S. Hassanpour and C. P. Langlotz, "Information extraction from multi-institutional radiology reports," *Artificial Intelligence in Medicine*, vol. 66, pp. 29–39, Jan. 2016, doi: 10.1016/j.artmed.2015.09.007.

[22] W. Yim, T. Denman, S. W. Kwan, and M. Yetisgen, "Tumor information extraction in radiology reports for hepatocellular carcinoma patients," *AMIA Jt Summits Transl Sci Proc*, vol. 2016, pp. 455–464, Jul. 2016.

[23] S. Cornegruta, R. Bakewell, S. Withey, and G. Montana, "Modelling Radiological Language with Bidirectional Long Short-Term Memory Networks," in *Proceedings of the Seventh International Workshop on Health Text Mining and Information Analysis*, Auxtin, TX, Nov. 2016, pp. 17–27. doi: 10.18653/v1/W16-6103.

[24] J. M. Steinkamp, C. Chambers, D. Lalevic, H. M. Zafar, and T. S. Cook, "Toward Complete Structured Information Extraction from Radiology Reports Using Machine Learning," *J Digit Imaging*, vol. 32, no. 4, pp. 554–564, Aug. 2019, doi: 10.1007/s10278-019-00234-y.

[25] K. Clark, U. Khandelwal, O. Levy, and C. D. Manning, "What Does BERT Look at? An Analysis of BERT's Attention," in *Proceedings of the 2019 ACL Workshop BlackboxNLP: Analyzing and Interpreting Neural Networks for NLP*, Florence, Italy, Aug. 2019, pp. 276–286. doi: 10.18653/v1/W19-4828.

[26] Y. Liu, F. Wei, S. Li, H. Ji, M. Zhou, and H. Wang, "A Dependency-Based Neural Network for Relation Classification," in *Proceedings of the 53rd Annual Meeting of the Association for Computational Linguistics and the 7th International Joint Conference on Natural Language Processing (Volume 2: Short Papers)*, Beijing, China, Jul. 2015, pp. 285–290. doi: 10.3115/v1/P15-2047.

[27] Y. Xu, L. Mou, G. Li, Y. Chen, H. Peng, and Z. Jin, "Classifying Relations via Long Short Term Memory Networks along Shortest Dependency Paths," in *Proceedings of the 2015 Conference on Empirical Methods in Natural Language Processing*, Lisbon, Portugal, Sep. 2015, pp. 1785–1794. doi: 10.18653/v1/D15-1206.

[28] K. Sugimoto *et al.*, "Extracting clinical terms from radiology reports with deep learning," *Journal of Biomedical Informatics*, vol. 116, p. 103729, Apr. 2021, doi: 10.1016/j.jbi.2021.103729.

[29] S. Miao *et al.*, "Extraction of BI-RADS findings from breast ultrasound reports in Chinese using deep learning approaches," *International Journal of Medical Informatics*, vol. 119, pp. 17–21, Nov. 2018, doi: 10.1016/j.ijmedinf.2018.08.009.

[30] X. Zhang *et al.*, "Extracting comprehensive clinical information for breast cancer using deep learning methods," *Int J Med Inform*, vol. 132, p. 103985, Dec. 2019, doi: 10.1016/j.ijmedinf.2019.103985.

[31] S. Datta, Y. Si, L. Rodriguez, S. E. Shooshan, D. Demner-Fushman, and K. Roberts, "Understanding spatial language in radiology: Representation framework, annotation, and spatial relation extraction from chest X-ray reports using deep learning," *Journal of Biomedical Informatics*, vol. 108, p. 103473, Aug. 2020, doi: 10.1016/j.jbi.2020.103473.

[32] K. Lybarger, M. Ostendorf, and M. Yetisgen, "Annotating social determinants of health using active learning, and characterizing determinants using neural event extraction," *J Biomed Inform*, vol. 113, p. 103631, Jan. 2021, doi: 10.1016/j.jbi.2020.103631.

[33] G. Hripcsak and A. S. Rothschild, "Agreement, the F-Measure, and Reliability in Information Retrieval," *J Am Med Inform Assoc*, vol. 12, no. 3, pp. 296–298, 2005, doi: 10.1197/jamia.M1733.

[34] G. Lample, M. Ballesteros, S. Subramanian, K. Kawakami, and C. Dyer, "Neural Architectures for Named Entity Recognition," in *Proceedings of the 2016 Conference of the North American Chapter of the Association for Computational Linguistics: Human Language Technologies*, San Diego, California, Jun. 2016, pp. 260–270. doi: 10.18653/v1/N16-1030.

[35] F. Dernoncourt, J. Y. Lee, and P. Szolovits, "NeuroNER: an easy-to-use program for named-entity recognition based on neural networks," in *Proceedings of the 2017 Conference on Empirical Methods in Natural Language Processing: System Demonstrations*, Copenhagen, Denmark, 2017, pp. 97–102. doi: 10.18653/v1/D17-2017.

[36] Y. Wu *et al.*, "Google's Neural Machine Translation System: Bridging the Gap between Human and Machine Translation," *arXiv:1609.08144 [cs]*, Sep. 2016, Accessed: May 15, 2019. [Online]. Available: http://arxiv.org/abs/1609.08144

[37] W. Lau, T. H. Payne, O. Uzuner, and M. Yetisgen, "Extraction and Analysis of Clinically Important Follow-up Recommendations in a Large Radiology Dataset," *AMIA Jt Summits Transl Sci Proc*, vol. 2020, pp. 335–344, 2020.

[38] L. Prechelt, "Automatic early stopping using cross validation: quantifying the criteria," *Neural Networks*, vol. 11, no. 4, pp. 761–767, Jun. 1998, doi: 10.1016/S0893-6080(98)00010-0.

[39] E. Alsentzer *et al.*, "Publicly Available Clinical BERT Embeddings," in *Proceedings of the 2nd Clinical Natural Language Processing Workshop*, Minneapolis, Minnesota, USA, Jun. 2019, pp. 72–78. doi: 10.18653/v1/W19-1909.

[40] A. E. W. Johnson *et al.*, "MIMIC-III, a freely accessible critical care database," *Scientific Data*, vol. 3, no. 1, pp. 1–9, May 2016, doi: 10.1038/sdata.2016.35.

[41] T. G. Dietterich, "Approximate Statistical Tests for Comparing Supervised Classification Learning Algorithms," *Neural Computation*, vol. 10, no. 7, pp. 1895–1923, Oct. 1998, doi: 10.1162/089976698300017197.

[42] C. Nadeau and Y. Bengio, "Inference for the Generalization Error," *Machine Learning*, vol. 52, no. 3, pp. 239–281, Sep. 2003, doi: 10.1023/A:1024068626366.

[43] G. Chauhan *et al.*, "Joint Modeling of Chest Radiographs and Radiology Reports for Pulmonary Edema Assessment," in *Medical Image Computing and Computer Assisted Intervention – MICCAI 2020*, Cham, 2020, pp. 529–539. doi: 10.1007/978-3-030-59713-9_51.

[44] J. P. Cohen *et al.*, "Predicting COVID-19 Pneumonia Severity on Chest X-ray With Deep Learning," *Cureus*, vol. 12, no. 7, p. e9448, doi: 10.7759/cureus.9448.

[45] E. Wu, K. Wu, R. Daneshjou, D. Ouyang, D. E. Ho, and J. Zou, "How medical AI devices are evaluated: limitations and recommendations from an analysis of FDA approvals," *Nat Med*, vol. 27, no. 4, pp. 582–584, Apr. 2021, doi: 10.1038/s41591-021-01312-x.

[46] D. Wadden, U. Wennberg, Y. Luan, and H. Hajishirzi, "Entity, Relation, and Event Extraction with Contextualized Span Representations," in *Proceedings of the 2019 Conference on Empirical Methods in Natural Language Processing and the 9th International Joint Conference on Natural Language Processing (EMNLP-IJCNLP)*, Hong Kong, China, Nov. 2019, pp. 5784–5789. doi: 10.18653/v1/D19-1585.

[47] Z. Zhong and D. Chen, "A Frustratingly Easy Approach for Entity and Relation Extraction," in *Proceedings of the 2021 Conference of the North American Chapter of the Association for Computational Linguistics: Human Language Technologies*, Online, Jun. 2021, pp. 50–61. doi: 10.18653/v1/2021.naacl-main.5.

[48] Y. Gu *et al.*, "Domain-Specific Language Model Pretraining for Biomedical Natural Language Processing," *arXiv:2007.15779 [cs]*, Sep. 2021, doi: 10.1145/3458754.